\documentclass[10pt,twocolumn]{article}
\usepackage{cvpr}
\usepackage{epsfig}

\def \ie {\emph{i.e.}\xspace}
\def \eg {\emph{e.g.}\xspace}
\def \etc {\emph{etc.}\xspace}
\def \etal {\emph{et al.}\xspace}

\newcommand{\para}[1]{\noindent \textbf{#1}}

\DeclareMathAlphabet{\mathpzc}{T1}{pzc}{m}{n}

\usepackage{amsmath}
\usepackage{tabularx} 
\usepackage{multirow} 
\usepackage{caption}

\usepackage[normalem]{ulem} 
\newcolumntype{C}[1]{>{\centering\let\newline\\\arraybackslash\hspace{0pt}}m{#1}} 
\makeatletter

\usepackage[pagebackref=true,breaklinks=true,colorlinks,bookmarks=false]{hyperref}

\cvprfinalcopy 

\begin{document}

\title{Crowd counting via scale-adaptive convolutional neural network}
\author{Lu Zhang\thanks{Equal contribution.}~$^\dag$\\
Tencent Youtu\\
{\tt\small v\_youyzhang@tencent.com}
\and
Miaojing Shi$^*$\\
Inria Rennes \& Tencent Youtu\\
{\tt\small miaojing.shi@inria.fr}
\and
Qiaobo Chen\thanks{Lu Zhang and Qiaobo Chen have worked on this paper during their internships at Tencent Youtu Lab.}\\
Shanghai Jiaotong University\\
{\tt\small chenqiaobo@sjtu.edu.cn}
}

\maketitle
\begin{abstract}
The task of crowd counting is to automatically estimate the pedestrian number in crowd images. To cope with the scale and perspective changes that commonly exist in crowd images, state-of-the-art approaches employ multi-column CNN architectures to regress density maps of crowd images. Multiple columns have different receptive fields corresponding to pedestrians (heads) of different scales. We instead propose a scale-adaptive CNN (SaCNN) architecture with a backbone of fixed small receptive fields. We extract feature maps from multiple layers and adapt them to have the same output size; we combine them to produce the final density map. The number of people is computed by integrating the density map.
We also introduce a relative count loss along with the density map loss to improve the network generalization on crowd scenes with few pedestrians, where most representative approaches perform poorly on. We conduct extensive experiments on the ShanghaiTech, UCF\_CC\_50 and WorldExpo'10 datasets as well as a new dataset SmartCity that we collect for crowd scenes with few people. The results demonstrate significant improvements of SaCNN over the state-of-the-art.

\end{abstract}

\section{Introduction}\label{Sec:Intro}
The crowd counting task in computer vision is to automatically count the pedestrian number in images/videos. To help with crowd control and public safety in many scenarios such as public rallies and sports, accurate crowd counting is demanded.

Early methods estimate the pedestrian number via detection, where each individual pedestrian in a crowd is detected by trained detectors~\cite{zhao2008pami,li2008icpr,ge2009cvpr,idrees2015pami}. This can be very hard if pedestrians are heavily occluded or densely spread. Present methods instead regress the pedestrian number on the whole image and achieve significant improvements~\cite{chan2008cvpr,chan2009iccv,liu2015iccv}. Handcrafted features like HOG~\cite{idrees2013cvpr} were soon outperformed by modern deep representations~\cite{onoro2016eccv,zhang2015cvpr}.

\begin{figure}[t]
	\centering
	\includegraphics[width=1\columnwidth]{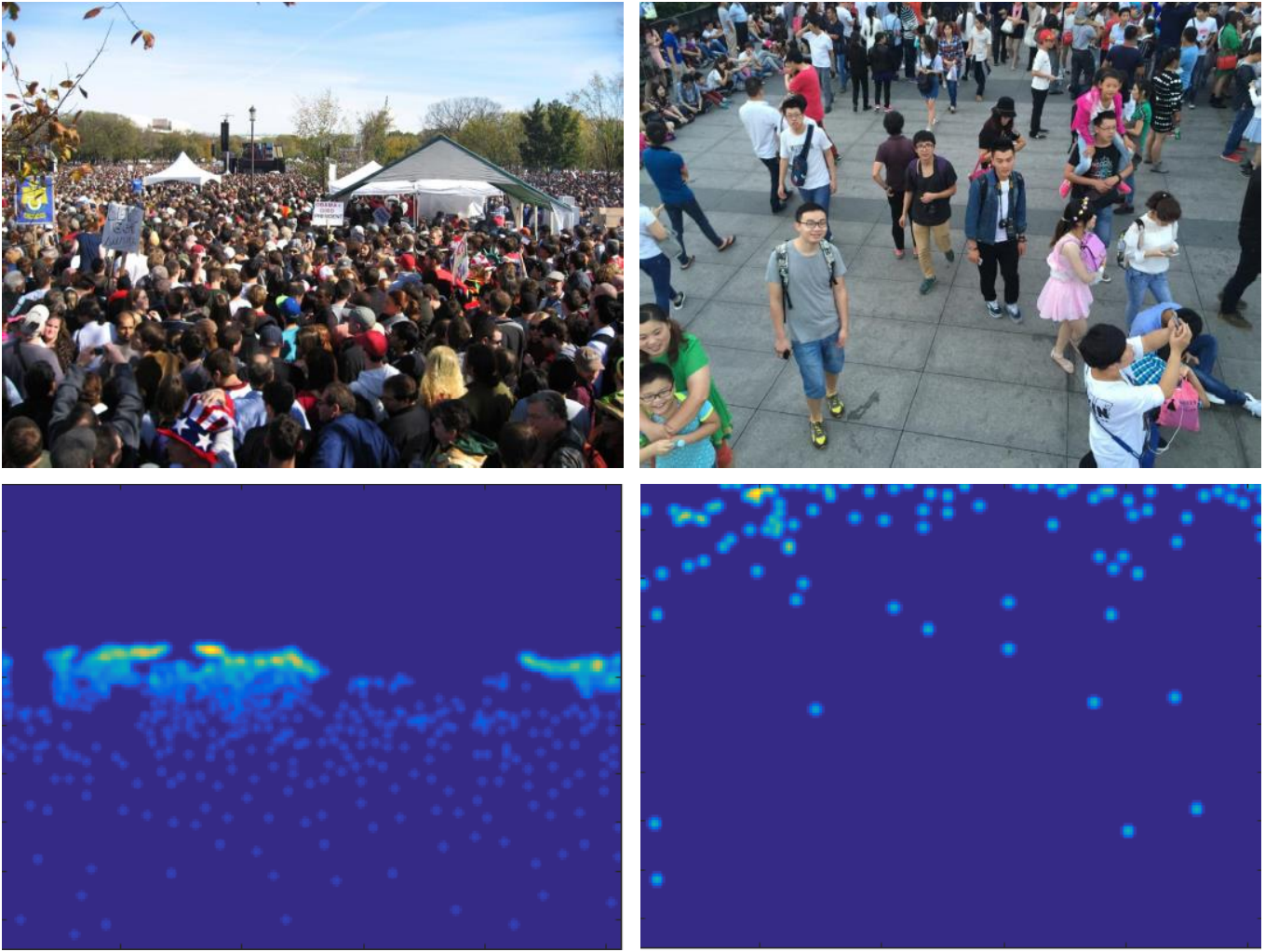}
	\caption{Top: crowd images. Bottom: ground truth density maps.}
	\label{Fig:densitymap}
\end{figure}

Crowd counting via regression suffers from the drastic changes in perspective and scale, which commonly exist in crowd images (see Fig.~\ref{Fig:densitymap}). To tackle it, multi-column convolutional neural networks (CNN) were recently adopted and have shown robust performance~\cite{sam2017arxiv,boominathan2016mm,zhang2016cvpr}. Different columns corresponding to different filter sizes (small, medium and large) are combined in the end to adapt to the large variations of perspective and scale. They regress a density map of the crowd (see Fig.~\ref{Fig:densitymap}). Pedestrian number is obtained by integrating the density map. \cite{sam2017arxiv} further introduced a switch classifier to relay the crowd patches in images to the best CNN column. Each CNN column is trained with its own samples.

We notice that the selection among the multiple columns in \cite{sam2017arxiv} is not balanced. Using one single column is able to retain over 70\% accuracy of the multi-column model on some datasets. Building upon this observation, in this paper, we propose a scale-adaptive CNN architecture (SaCNN, see Fig.~\ref{Fig:network}) for crowd counting. It offers several new elements as our contributions:  (1) We design a single column CNN with a single filter size as our backbone, which is easy to train from scratch. Small-sized filters preserve the spatial resolution and allow us to build a deep network. (2) We combine the feature maps of multiple layers to adapt the network to the changes in pedestrian (head) scale and perspective. Different layers share the same low-level  feature representations, which results in fewer parameters, fewer training data required, and faster training.
(3) We introduce a multi-task loss by adding a relative head count loss to the density map loss. It significantly improves the network generalization on crowd scenes with few pedestrians, where most representative works perform poorly on. (4) We collect the new SmartCity dataset with high-angle shot for crowd counting. The existing datasets are taken from outdoors and do not adequately cover those crowd scenes with few pedestrians. The new dataset contains both indoor and outdoor scenes and has an average pedestrian number of 7.4 per image.
(5) We conduct extensive experiments on three datasets: ShanghaiTech~\cite{zhang2016cvpr}, UCF\_CC\_50~\cite{idrees2013cvpr} and WorldExpo'10~\cite{zhang2015cvpr}; the results show that our SaCNN significantly outperforms the state-of-the-art crowd counting methods. We also compare SaCNN with other representative works on SmartCity to show the generalization ability of ours. Our code and dataset are publicly available\footnote{\footnotesize\color{red} http://github.com/miao0913/SaCNN-CrowdCounting-Tencent\_Youtu}.

\section{Related work}
We categorize crowd counting methods as either 1) detection-based methods or 2) regression-based methods. We shall describe both categories below and compare representative works to our method.

\subsection{Detection-based methods}
Detection-based methods consider a crowd as a group of detected individual entities~\cite{wang2011cvpr,wu2005iccv,stewart2016cvpr,viola2003ijcv,brostow2006cvpr,rabaud2006cvpr}.
Early works focus on video surveillance scenario so that the motion and appearance cues can be utilized~\cite{viola2003ijcv,brostow2006cvpr,rabaud2006cvpr}. For example, \cite{viola2003ijcv} trained a dynamic detector capturing motion and appearance information over two consecutive frame pairs of a video sequence.
These works are not applicable in still images for crowd counting. Pedestrians are often occluded in dense crowds, and this is particularly challenging when estimating the crowds in still images. Part-detectors were therefore employed to count pedestrian from parts in images~\cite{wang2011cvpr,wu2005iccv}. \cite{stewart2016cvpr} took advantage of a recurrent deep network to detect the occluded heads in crowd scenes. Notwithstanding their improvements, detection-based methods overall suffer severely in dense crowds with high occlusions among people.

\subsection{Regression-based methods}
Regression-based methods regress either a scalar value (pedestrian number)~\cite{chan2008cvpr,chan2009iccv,liu2015iccv,idrees2013cvpr} or a density map~\cite{chen2012bmvc,kong2005bmvc,lempitsky2010nips} on various features extracted from crowd images. They basically have two steps: first, extracting effective features from crowd images; second, utilizing various regression functions to estimate the crowd count. Early works utilize handcrafted features, like edge features~\cite{chan2008cvpr,chen2012bmvc,ryan2009dicta,regazzoni1996sp} and texture features~\cite{chen2012bmvc,idrees2013cvpr,marana1998sibgrapi}. Regression methods include linear~\cite{regazzoni1996sp,paragios2001cvpr}, ridge~\cite{chen2012bmvc} and Gaussian~\cite{chan2008cvpr} functions.

Due to the use of strong CNN features, recent works on crowd counting have shown remarkable progress by regressing a density map of an image~\cite{zhang2015cvpr,zhang2016cvpr,sam2017arxiv,boominathan2016mm,onoro2016eccv}. 
A density map provides much more information than a scalar value, and CNNs have been demonstrated to be particularly good at solving problems ``locally" rather than ``globally". 
Crowd counting is casted as that of estimating a continuous density function whose integral over any image region gives the count of pedestrians within that region.
\cite{zhang2015cvpr} designed a multi-task network to regress both density maps and crowd counts. They employed fully-connected layers to regress the absolute crowd count. 

\begin{figure*}[t]
	\centering
	\includegraphics[width=1\textwidth]{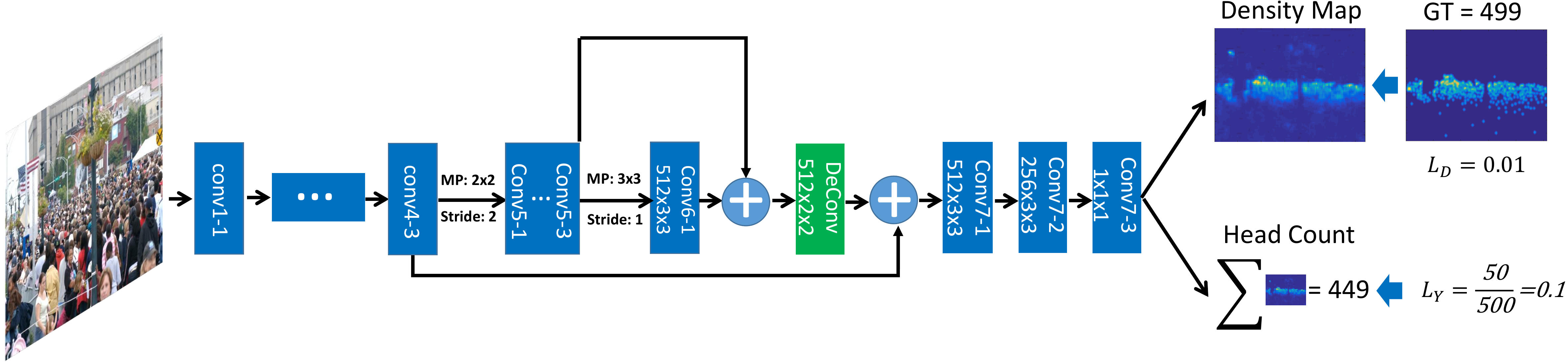}
	\caption{The structure of the proposed scale-adaptive convolutional neural network (SaCNN) for crowd counting. MP: max pooling layer; DeConv: deconvolutional layer; GT: ground truth.
We build a single backbone network with a single filter size.
We combine the feature maps of multiple layers to adapt the network to the changes in pedestrian scale and perspective. Multi-scale layers share the same low-level parameters and feature representations, which results in fewer parameters, fewer training data required, and faster training. We introduce two loss
functions to jointly optimize the model: one is density map loss, the other is relative count loss. The relative count loss helps to reduce the variance of the prediction errors and improve the network generalization on very sparse crowd scenes.
}
	\label{Fig:network}
\end{figure*}
\medskip

\subsection{Comparison to our SaCNN}
\cite{zhang2016cvpr,sam2017arxiv,boominathan2016mm} utilized multi-column networks~\cite{ciregan2012cvpr} to deal with the variation of head scale in one image. Different columns correspond to different filter sizes. The large-sized filters make the whole network hard to train. The network is pretrained on each column first and finetuned together in the end~\cite{zhang2016cvpr}; whilst our SaCNN employs a single column with a single filter size which can be trained from scratch. \cite{sam2017arxiv} introduced a switch classifier using VGG net~\cite{simonyan2015iclr} to relay the crowd patches from images to the best CNN column. The switching is very costly and often not correct. While our SaCNN adapts the feature maps from multiple scales and concatenates them to relay each pedestrian to the best scale in the network.

Similar to~\cite{zhang2015cvpr}, we perform a multi-task optimization in SaCNN to regress both the density map and head count. Differently, ours is a multi-scale fully convolutional network and we propose a relative count loss. The scale-adaptive architecture is similar in spirit to~\cite{ronneberger2015miccai,lin2017cvpr} by combining feature maps of multiple layers; but we use the deconvolutional layer~\cite{noh2015iccv} to adapt the network output instead of upsampling and element-wise summation. Our network design with a backbone of single filter size and multi-scale outputs is new in the crowd counting field. Plus, we claim five-fold new elements in total in Sec.~\ref{Sec:Intro}.



\section{Scale-adaptive CNN}\label{Sec:Method}
In this section, we first generate the ground truth density maps for training data and then present the architecture of SaCNN.
\subsection{Ground truth density maps}
Supposing we have a pedestrian head at a pixel $z_j$, we represent it by a delta function $\delta(z - z_j)$. The ground truth density map $D(z)$ is computed by convolving this delta function using a Gaussian kernel ${G_{{\sigma}}}$ normalized to sum to one:
\begin{equation}\label{Eq:densitymap}
 D(z) = \sum\limits_{j = 1}^M {\delta (z - {z_j}) * {G_{{\sigma}}}(z)},
\end{equation}
where $M$ is the total head number. The sum of the density map is equivalent to the total number of pedestrians in an image.
Parameter setting follows~\cite{zhang2016cvpr,sam2017arxiv,boominathan2016mm}.



\subsection{Network architecture}\label{Sec:Scale}
Many state-of-the-art works~\cite{zhang2016cvpr,sam2017arxiv,boominathan2016mm}  adopt the multi-column architecture with different filter sizes to address the scale and perspective change in crowd images. We instead propose a scale-adaptive CNN (SaCNN) where we use a single backbone network with a single filter size (see Fig.~\ref{Fig:network}). Density map estimation is performed at a concatenated layer which merges the feature maps from multiple scales of the network; it can therefore easily adapt to the pedestrian scale and perspective variations.

We use all $3*3$ filters in the network. The computation required for $3*3$ filters is far less than for large filters~\cite{zhang2016cvpr,sam2017arxiv,boominathan2016mm}. Small filters preserve the spatial resolution of the input so that we can build a deep network.
``Deeper" has a similar effect as ``Wider" (multi-column architecture) does in a network: for instance, a large filter can be simulated by a small filter from deeper layer.
The backbone design (until conv4\_3 in Fig.~\ref{Fig:network}) follows the VGG architecture~\cite{simonyan2015iclr} in the context of using 3*3 filters and three convolutional layers after each pooling layer.

\begin{figure}[t]
	\centering
	\includegraphics[width=1\columnwidth]{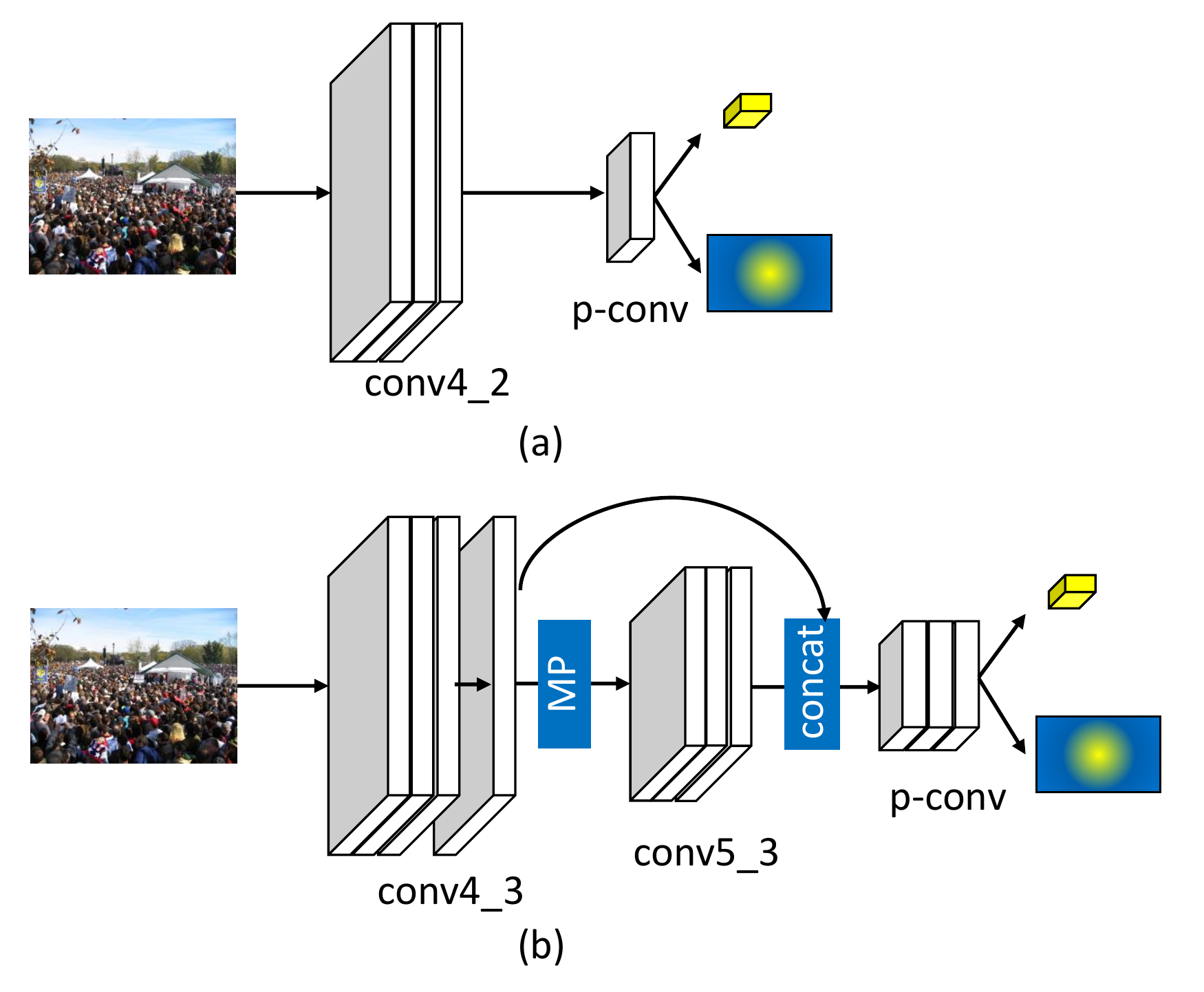}
	\caption{(a) single-scale model; (b) two-scale model. MP: max pooling layer of stride 1. The outputs of the networks are the density maps (blue-yellow heatmaps) and head counts (yellow cubes). p-conv denotes $1\times1$ convolutional layer.  }
	\label{Fig:depth}
\end{figure}

SaCNN is built from low-level to high-level by gradually adding layers of different scales.
For concatenation, we need to carefully adapt their output to be the same size. We first present a \emph{single-scale} model that is illustrated in
Fig.~\ref{Fig:depth} (a): it takes the output feature map from conv4\_2 and projects it to a density map via $1 \times 1$ filters in p-conv. Results in Fig.~\ref{Fig:multiscale} show that our single-scale model converges fast and its performance is close to the state-of-the art.

Like the multi-column CNN, we want our model to fire on pedestrians (heads) of different sizes. We decide to go deeper. We have three max-pooling layers and several convolutional layers until conv4\_3. We add pool4 and three convolutional layers (conv5\_1 to conv5\_3) after conv4\_3 in the network (see Fig.~\ref{Fig:depth} (b)). We propose a \emph{two-scale} model by extracting feature maps from conv4\_3 and conv5\_3; we set the stride of pool4 to be 1 so that the two feature maps have the same size; we concatenate them to produce the final density map (refer to Fig.~\ref{Fig:depth} (b)) . Results in Fig.~\ref{Fig:multiscale} demonstrate a clear improvement of our two-scale model over the single-scale model.


Building upon the two-scale network, we construct our \emph{scale-adaptive} model in Fig.~\ref{Fig:network}.
We add pool5 and conv6\_1 afterwards in the network. In order to combine conv5\_3 and conv6\_1, we set the stride of pool4 as 2 and pool5 as 1. We upsample the concatenated output to $1/8$ resolution of the input
using DeConv layer to further concatenate it with conv4\_3.
The final density map therefore has a spatial resolution of $1/8$ times the input Hence, we downsample the ground truth density map by factor 8.



Notice that we have also tried to regress the density map at a lower resolution: \eg $1/16$ times the input resolution.
The performance is slightly declined compared to concatenating at $1/8$ resolution. We suggest it is due to the receptive field being too big in the deeper layer.

\subsection{Network loss}
Our network training first adopts the Euclidean loss to measure the distance between the estimated density map and the ground truth~\cite{zhang2016cvpr,sam2017arxiv,boominathan2016mm}:
\begin{equation}\label{eq_density}
  {L_D}(\Theta ) = \frac{1}{{N}}\sum\limits_{i = 1}^N {\|{F_D}({X_i};\Theta ) - {D_i}\|^2},
\end{equation}
where $\Theta$ is a set of learnable parameters in the proposed network. $N$ is the number of training images. $X_i$ is the input image and $D_i$ is the corresponding ground truth density map. ${{F_D}({X_i};\Theta )}$ denotes the estimated density map for $X_i$. The Euclidean distance is computed at pixels and summed over them.

Apart from the density map regression, we introduce another loss function regarding the head count (see Fig,~\ref{Fig:network}). We notice that most representative approaches perform poorly on crowd scenes with few pedestrians. This problem can not be resolved via (\ref{eq_density}), because the absolute pedestrian number is usually not very large in sparse crowds compared to that in dense crowds.
To tackle this, we propose a relative head count loss:
\begin{equation}\label{eq_count}
  {L_Y}(\Theta ) = \frac{1}{{N}}\sum\limits_{i = 1}^N {\|\frac{{{F_Y}({X_i};\Theta ) - {Y_i}}}{{{Y_i +1}}}\|^2},
\end{equation}
where ${{F_Y}({X_i};\Theta )}$ and $Y_i$ are the estimated head count and the ground truth head count, respectively. The denominator $Y_i$ is added by $1$ to prevent division by zero. (\ref{eq_count}) concentrates the learning on those samples with relatively large prediction errors. Results on those very sparse crowd scenes (Sec.~\ref{Sec:Experiment}: Table~\ref{Tab:SmartCity}) demonstrate significant improvements by employing the relative head count loss in the network.

\medskip

We directly train our SaCNN from scratch by randomly initializing the network parameters. Because the head count regression is an easy task, we first regress the model on the density map (\ref{eq_density}); once it converges, we add (\ref{eq_count}) in the objective loss to jointly train a multi-task network for a few more epochs. We set the density map loss weight as 1 and the relative count loss weight as 0.1.

We note that it will be 1.5$\times$ faster for the model to converge if we train a two-scale model first and then finetune SaCNN on it; the performance difference between the two training manners is trivial.

\section{Experiments}\label{Sec:Experiment}
We shall first briefly introduce three standard datasets and a new created dataset for crowd counting. Afterwards, we evaluate our method on these datasets.
\begin{table*}
	\centering
	\small
\begin{tabular}{|l|c|c|c|c|c|c|c|}
          \hline
	 \multicolumn{2}{|c|}{Dataset}&	Resolution	& \#Images & Min & Max & Avg & Total\\
          \hline
	 \hline
      \multirow{2}{*}{ShanghaiTech}& PartA & different & 482 &  33& 3139 & 501.4 & 241,677  \\
      \cline{2-8}
      & PartB& $768*1024$ & 716 & 9& 578 & 123.6 & 88,488\\
      \hline
       \multicolumn{2}{|c|}{WorldExpo'10} &$576*720$ & 3980 &1 & 253 & 50.2 & 199,923 \\
       \hline
        \multicolumn{2}{|c|}{UCF\_CC\_50} & different& 50& 94 & 4543 & 1279.5 & 63,974\\
        \hline
          \multicolumn{2}{|c|}{SmartCity} &1920*1080& 50 & 1 & 14 & 7.4 & 369\\
           \hline
\end{tabular}
	\caption{Dataset statistics in this paper. \#Images is the number of images; Min, Max and Avg denote the minimum, maximum, and average pedestrian numbers per image, respectively. }
	\label{Tab:Statis}
\end{table*}
%
\subsection{Datasets}
\para{ShanghaiTech dataset~\cite{zhang2016cvpr}} consists of 1198 annotated images with a total of 330,165 people with head center annotations. This dataset is split into two parts: PartA and PartB. Refer to Table~\ref{Tab:Statis} for the statistics of the dataset: the crowd images are sparser in PartB compared with PartA. Following~\cite{zhang2016cvpr}, we use 300 images for training and 182 images for testing in PartA; 400 images for training and 316 images for testing in PartB.
\medskip

\para{WorldExpo'10 dataset~\cite{zhang2015cvpr}} includes 3980 frames, which are from the Shanghai 2010 WorldExpo. 3380 frames are used as training while the rest are taken as test. The test set includes five different scenes and 120 frames in each one. Regions of interest (ROI) are provided in each scene so that crowd counting is only conducted in the ROI in each frame. Some statistics of this dataset
can be found in Table~\ref{Tab:Statis}.
\medskip

\para{UCF\_CC\_50 dataset~\cite{idrees2013cvpr}} has 50 images with 63974 head annotations in total. The head counts range between 94 and 4543 per image. The small dataset size and large variance in crowd count make it a very challenging dataset. Following~\cite{idrees2013cvpr}, we perform 5-fold cross validations to report the average test performance. Some statistics of the dataset is in Table~\ref{Tab:Statis}.
\medskip

\begin{figure}[t]
	\centering
	\includegraphics[width=1\columnwidth]{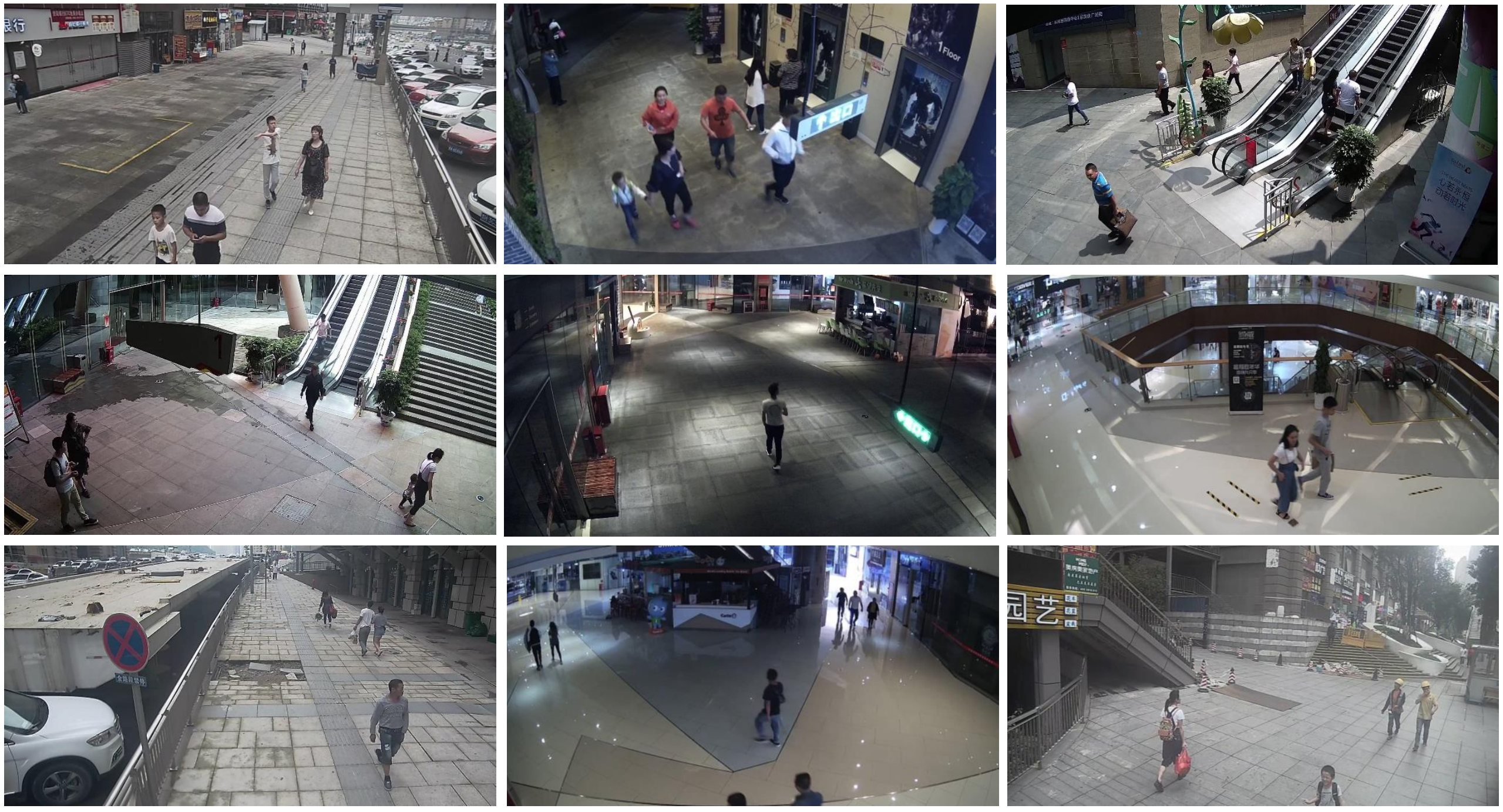}
	\caption{Examples for the SmartCity dataset. They consist of both indoor and outdoor scenes with few pedestrians. }
	\label{Fig:smartcity}
\end{figure}

\para{SmartCity dataset} is collected by ourselves. There are in total 50 images collected from ten city scenes including office entrance, sidewalk, atrium, shopping mall \etc. Some examples are shown in Fig.~\ref{Fig:smartcity}. They are all high-angle shot for video surveillance. Existing crowd counting datasets consist of images of hundreds or even thousands of pedestrians, and nearly all the images are taken outdoors. We therefore specifically create the dataset that has few pedestrians in images and consists of both outdoor and indoor scenes. Some statistics of this dataset is shown in Table~\ref{Tab:Statis}: the average number of pedestrians per image is only 7.4 with minimum being 1 and maximum being 14. We use this dataset to test the generalization ability of the proposed framework on very sparse crowd scenes.
\medskip

\begin{figure*}[t]
	\centering
	\includegraphics[width=1\textwidth]{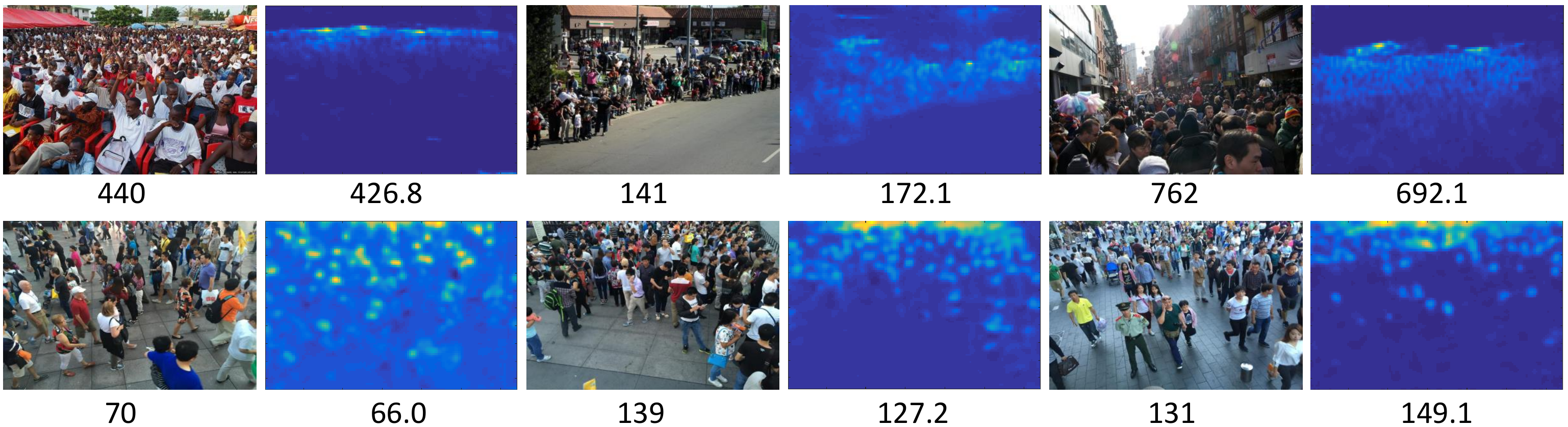}
	\caption{Result on ShanghaiTech dataset. Top: PartA; Bottom: PartB. We present six test images and their estimated density maps on the right. Ground truth and the estimated pedestrian numbers are beneath the real images and the corresponding density maps, respectively. }
	\label{Fig:shanghaitech}
\end{figure*}

\subsection{Implementation details and evaluation protocol}
Ground truth annotations for each head center in the standard benchmarks are publicly available. Given a training set, we augment it by randomly cropping 9 patches from each image. Each patch is $1/4$ size of the original image. All patches are used to train our model SaCNN.
We train the model using the stochastic gradient descent (SGD) optimizer. The learning rate starts from 1e-6 and decays to 1e-8 with multistep policy. Momentum is 0.9 and batch size is 1, we train 250 epochs in total.

We evaluate the performance via the mean absolute error (MAE) and mean square error (MSE) commonly used in previous works~\cite{zhang2015cvpr,zhang2016cvpr,sam2017arxiv,boominathan2016mm,onoro2016eccv,wang2015mm}:
\begin{equation}\label{Eq:MAE}
\begin{split}
   &\mathrm{MAE} = \frac{1}{{N}}\sum\limits_{i = 1}^N {|{{F_Y} - {Y_i}}|},\\
   &\mathrm{MSE} = \sqrt{\frac{1}{{N}}\sum\limits_{i = 1}^N {({{F_Y} - {Y_i}})^2}}.
\end{split}
\end{equation}
One can refer to~(\ref{eq_count}) for notations $F_Y$ and $Y_i$. Small MAE and MSE indicate good performance.

\begin{figure}[t]
	\centering
	\includegraphics[width=1\columnwidth]{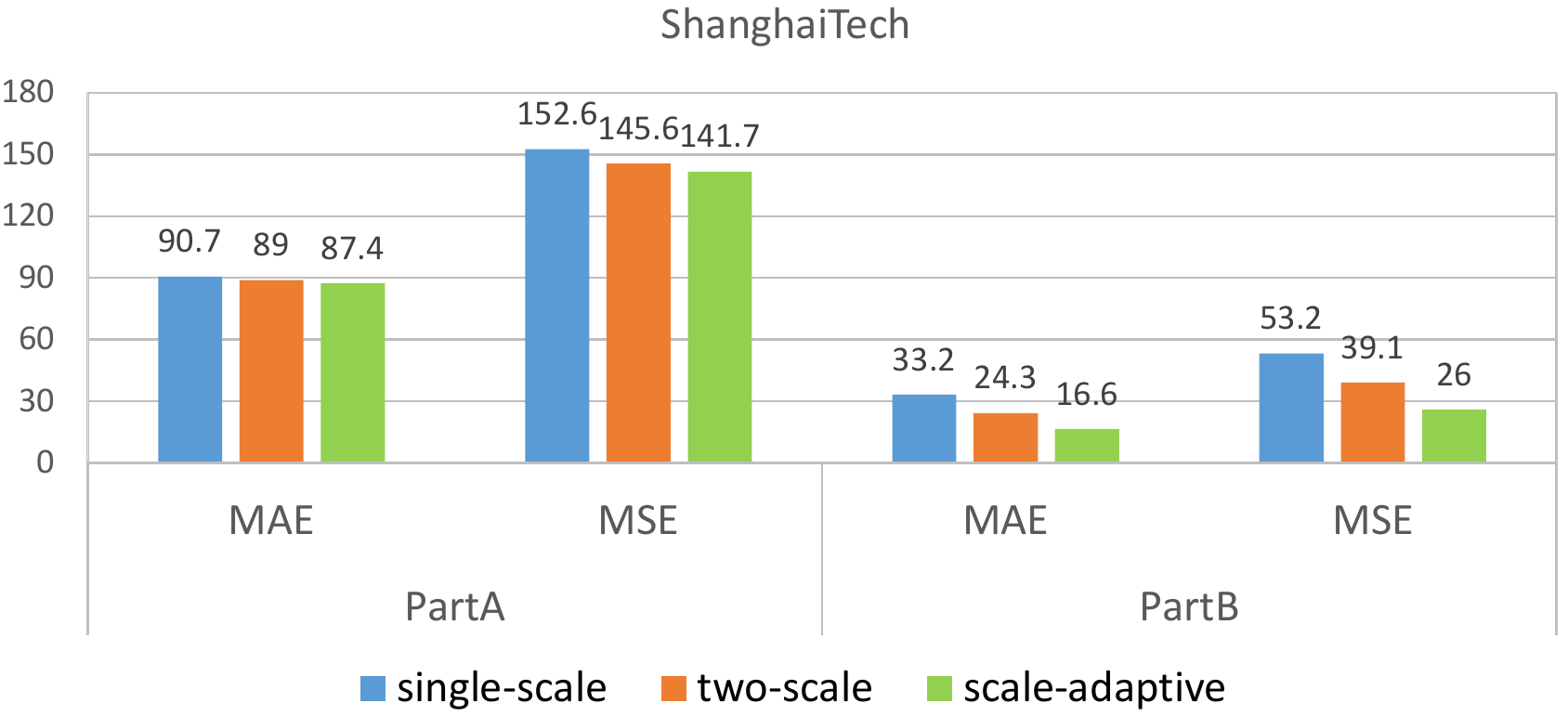}
	\caption{Multi-scale ablation test on ShanghaiTech dataset. The MAE and MSE are clearly decreased from single-scale to two-scale and scale-adaptive model.}
	\label{Fig:multiscale}
\end{figure}

\subsection{Results on ShanghaiTech}\label{Sec:ShanghaiTech}
\para{Ablation study.} We report an ablation study to offer the justification of our SaCNN in multi-scale and multi-task setup. Referring to Sec.~\ref{Sec:Method}, SaCNN is built from low-level to high-level by gradually adding layers of different scales. Network training is at first focused on the density map loss, after it converges, the relative count loss is added to continue the training.

\emph{Multi-scale ablation test.} We train three models separately corresponding to single-scale, two-scale, and three-scale (scale-adaptive) CNNs in Sec.~\ref{Sec:Method}. Fig.~\ref{Fig:multiscale} illustrates their performance. It can be seen that the MAE and MSE are clearly decreased from single-scale to two-scale models; eventually scale-adaptive model reaches the lowest MAE and MSE on both PartA and PartB. There are on average 4 times more people in PartA than in PartB, pedestrian heads are therefore quite small; whilst the feature maps from the deep layers tend to fire on big heads (Sec.~\ref{Sec:Scale}). Thus combining multi-scale outputs does not result in a significant decrease in MAE and MSE on PartA.
But still, the results validate our argument that the complementary scale-specific feature maps produce a strong scale-adaptive crowd counter.
We use the three-scale model in the following experiments. Notice in this test we only employ the density map loss during training.

\emph{Multi-task ablation test.} We introduce a relative count loss (rcl) to the above scale-adaptive model to improve the network generalization ability on crowd scenes with few pedestrians. We denote by SaCNN(w/o cl) the scale-adaptive model without count loss and SaCNN(rcl) with relative count loss. It can be seen from Table~\ref{Tab:Alation} that, compared to SaCNN(w/o cl), SaCNN(rcl) decreases both MAE and MSE on PartA and PartB.


To justify the relative count loss over the absolute count loss (acl), we train another model SaCNN(acl) following~\cite{zhang2015cvpr}, where they train a multi-task network by alternating between training on the density map loss and the absolute count loss. The result in Table.~\ref{Tab:Alation} shows that adding the absolute loss however impairs the performance. This is because the absolute count loss varies drastically among crowd images. In the following experiments, we use SaCNN to signify our best model SaCNN(rcl). We shall also show the significant improvement of adding rcl on SmartCity dataset (Sec.~\ref{Sec:SmartCity}).

\begin{table}
	\setlength{\tabcolsep}{2.6pt}
	\centering
	\small
\begin{tabular}{|c|c|c|c|c|}
          \hline
    ShanghaiTech  & \multicolumn{2}{c|}{PartA}& \multicolumn{2}{c|}{PartB} \\
      \hline
     Method & MAE& MSE & MAE & MSE \\
     \hline
    SaCNN(w/o cl)&87.4	& 141.7&	16.6&	26.0\\
    SaCNN(acl)  & 102.7 & 164.9 & 17.5& 26.5 \\
    SaCNN(rcl)  & \textbf{86.8} & \textbf{139.2 }& \textbf{16.2}& \textbf{25.8} \\
     \hline
\end{tabular}
	\caption{Multi-task ablation test of SaCNN on ShanghaiTech dataset. cl is short for count loss; acl denotes the absolute count loss while rcl denotes the relative count loss proposed in this paper. }
	\label{Tab:Alation}
\end{table}

\begin{table}
	\setlength{\tabcolsep}{2.6pt}
	\centering
	\small
\begin{tabular}{|c|c|c|c|c|}
          \hline
    ShanghaiTech  & \multicolumn{2}{c|}{PartA}& \multicolumn{2}{c|}{PartB} \\
      \hline
     Method & MAE& MSE & MAE & MSE \\
     \hline
     Zhang~\etal~\cite{zhang2015cvpr} & 181.8 & 277.7 & 32.0&49.8 \\
     \hline
     Zhang~\etal~\cite{zhang2016cvpr} & 110.2 & 173.2& 26.4& 41.3 \\
    \hline
     Sam~\etal~\cite{sam2017arxiv}  & 90.4 & \textbf{135.0} & 21.6 & 33.4 \\
     \hline
     SaCNN    & \textbf{86.8} & 139.2 & \textbf{16.2} & \textbf{25.8} \\
     \hline
\end{tabular}
	\caption{Comparison of SaCNN with other state-of-the-art on ShanghaiTech dataset.}
	\label{Tab:Shanghaitech}
\end{table}
\medskip

\para{Comparison to state-of-the-art.} We compare our best model SaCNN with state-of-the-art~\cite{zhang2015cvpr,zhang2016cvpr,sam2017arxiv} on both PartA and PartB (Table~\ref{Tab:Shanghaitech}).
Our method achieves the best MAE: 86.8 and 16.2 on PartA and PartB. Compared to~\cite{sam2017arxiv}, SaCNN achieves 3.6 point decrease in MAE on PartA and 5.4 point on PartB; 7.6 point decrease in MSE on PartB and a comparable MSE (second best) on PartA. SaCNN requires less computation compared to~\cite{sam2017arxiv}, where
they split the test image into patches and employ both a classification and a regression network to test each patch.
At test time SaCNN is 2 times faster than~\cite{sam2017arxiv} (293ms vs. 580ms per image).

Fig.~\ref{Fig:shanghaitech} shows some examples on both PartA and PartB. The estimated density maps are visually similar to the crowd distributions in real images. We give the predicted number and the real pedestrian number under the density maps and real images, respectively. The estimated pedestrian numbers are close to the real numbers.

\subsection{Results on WorldExpo'10}
Referring to~\cite{zhang2015cvpr}, training and test are both conducted within the ROI provided for each scene. MAE is reported for each test scene and averaged to evaluate the overall performance. We compare our SaCNN with other state-of-the-art in Table~\ref{Tab:WorldExpo}.
Notations \ie (GT w/o perspective) and (GT with perspective) signify the selection of $\sigma$ in (\ref{Eq:densitymap}).
\cite{zhang2015cvpr,zhang2016cvpr} compute the $\sigma$ as a ratio of the perspective value at certain pixel. \cite{sam2017arxiv} however reports a better result without using perspective values.
We provide both results with and w/o perspectives. It can be seen that except on S4, SaCNN (GT with perspective) is inferior to SaCNN (GT w/o perspective) on all the remaining scenes. SaCNN(GT w/o perspective) produces the best MAE on S1, S2 and S5: 2.6, 13.5 and 3.3. The average MAE of SaCNN across scenes outperforms~\cite{sam2017arxiv} by 0.9 point and reaches the best 8.5.

\begin{table}
	\setlength{\tabcolsep}{1pt}
	\centering
	\small
\begin{tabular}{|c|c|c|c|c|c|c|}
          \hline
     WorldExpo'10 & S1& S2 & S3 & S4 & S5 & Avg.\\
     \hline
     Zhang~\etal~\cite{zhang2015cvpr}& 9.8 & 14.1 & 14.3& 22.2 & 3.7 & 12.9\\
    Zhang~\etal~\cite{zhang2016cvpr}  & 3.4 & 20.6 & 12.9 & 13.0 & 8.1 & 11.6 \\
     Sam~\etal~\cite{sam2017arxiv}   &4.4  & 15.7 & \textbf{10.0} & 11.0 & 5.9 & 9.4 \\
    SaCNN (GT with perspective)  & 4.4& 26.3 & 16.8 & \textbf{9.0} & 4.9 & 12.3 \\
    SaCNN (GT w/o perspective)  & \textbf{2.6}& \textbf{13.5} & 10.6 & 12.5 & \textbf{3.3}& \textbf{8.5} \\
     \hline
\end{tabular}
	\caption{Comparison of SaCNN with other state-of-the-art on WorldExpo'10 dataset. MAE is reported for each test scene and averaged in the end.}
	\label{Tab:WorldExpo}
\end{table}

\subsection{Results on UCF\_CC\_50}
We compare our method with seven existing methods on UCF\_CC\_50~\cite{zhang2015cvpr,zhang2016cvpr,sam2017arxiv,lempitsky2010nips,idrees2013cvpr,boominathan2016mm,onoro2016eccv} in Table~\ref{Tab:UCF}. \cite{lempitsky2010nips} proposed a density regression model to regress the density map rather than the pedestrian number in crowd counting. \cite{idrees2013cvpr} used multi-source features to estimate crowd count while \cite{boominathan2016mm} used a multi-column CNN with one column initialized by VGG net~\cite{simonyan2015iclr}. \cite{onoro2016eccv} adopted a custom CNN network trained separately on each scale; fully connected layers are used to fuse the maps from each of the CNN trained on a particular scale.

Our method achieves the best MAE 314.9 and MSE 424.8 compared to the state-of-the-art. The smallest MSE indicating the lowest variance of our prediction across the dataset.

\begin{table}
	\setlength{\tabcolsep}{2.6pt}
	\centering
	\small
\begin{tabular}{|c|c|c|}
          \hline
     UCF\_CC\_50 & MAE& MSE\\
     \hline
        Lempitsky~\etal~\cite{lempitsky2010nips}& 493.4 & 487.1 \\
        Idrees~\etal~\cite{idrees2013cvpr}& 419.5 & 541.6 \\
     Zhang~\etal~\cite{zhang2015cvpr}& 467.0 & 498.5 \\
       Boominathan~\etal~\cite{boominathan2016mm}& 452.5 & - \\
       Zhang~\etal~\cite{zhang2016cvpr}  & 377.6 & 509.1 \\
        Onoro~\etal~\cite{onoro2016eccv}& 333.7 & 425.3 \\
     Sam~\etal~\cite{sam2017arxiv}   & 318.1 & 439.2 \\
    SaCNN  & \textbf{314.9}& \textbf{424.8} \\
     \hline
\end{tabular}
	\caption{Comparison of SaCNN with other state-of-the-art on UCF\_CC\_50 dataset.}
	\label{Tab:UCF}
\end{table}

\begin{figure}[t]
	\centering
	\includegraphics[width=1\columnwidth]{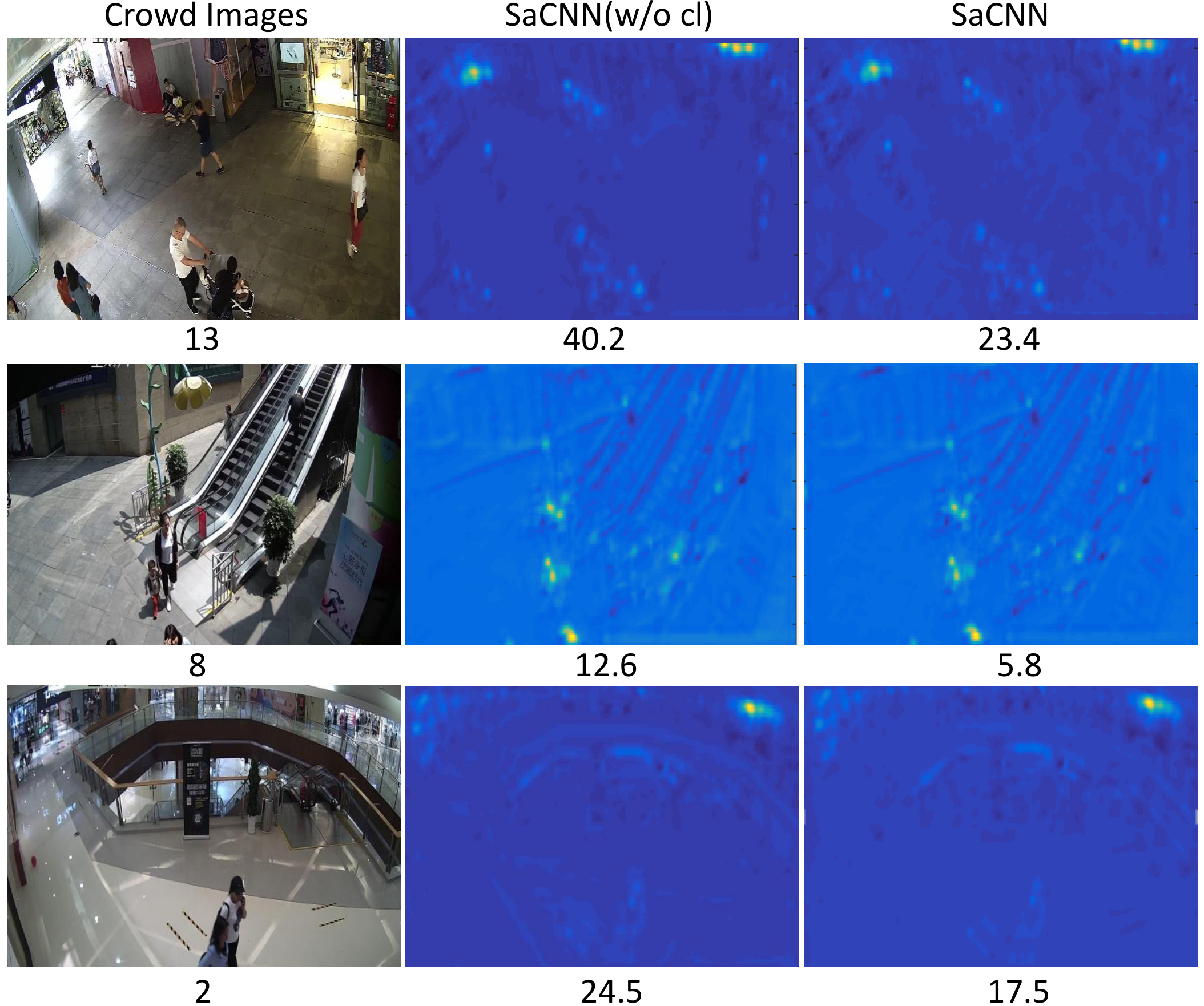}
	\caption{Comparison of SaCNN and SaCNN(w/o cl). The numbers below the real images are the ground truth. The second column denotes the density maps produced by SaCNN(w/o cl) while the third column is produced by SaCNN. Numbers below the density maps are the corresponding estimated pedestrian counts. }
	\label{Fig:smartcityres}
\end{figure}

\begin{table}
	\setlength{\tabcolsep}{2.6pt}
	\centering
	\small
\begin{tabular}{|c|c|c|}
          \hline
     SmartCity & MAE& MSE\\
     \hline
          Zhang~\etal~\cite{zhang2016cvpr}  & 40.0 & 46.2 \\
          Sam~\etal~\cite{sam2017arxiv}& 23.4 & 25.2 \\
    SaCNN(w/o cl)  & 17.8 & 23.4 \\
     SaCNN  & \textbf{8.6}& \textbf{11.6} \\
     \hline
\end{tabular}
	\caption{Comparison of SaCNN with other representative works on SmartCity dataset.}
	\label{Tab:SmartCity}
\end{table}

\begin{figure*}[t]
	\centering
	\includegraphics[width=1\textwidth]{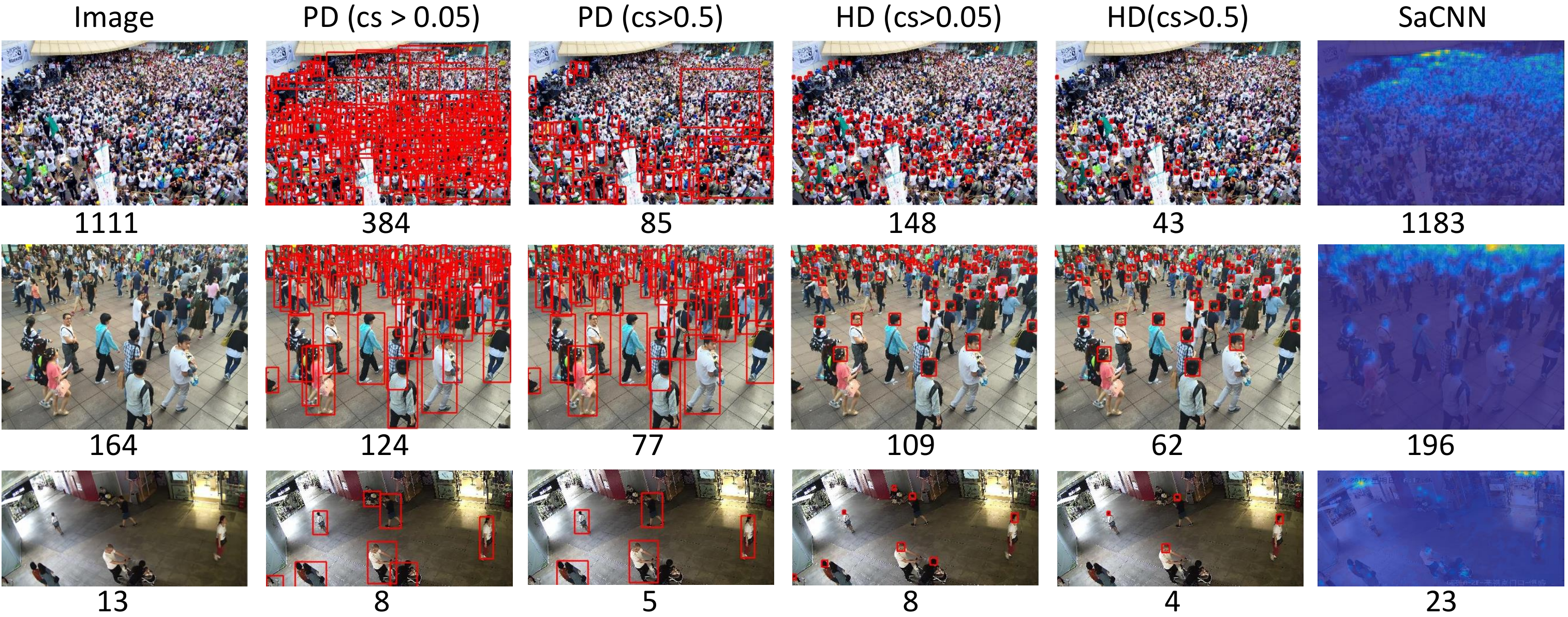}
	\caption{Comparison between YOLO9000~\cite{redmon2017cvpr} and SaCNN. PD: pedestrian detector; HD: head detector; $cs$ confidence score. The numbers below the real images (first column) are the ground truth.  Numbers below other images are the estimated pedestrian counts.}
	\label{Fig:detection}
\end{figure*}

\begin{table*}
	\centering
	\small
\begin{tabular}{|c|c|c|c|c|c|c|}
          \hline
        Dataset  & \multicolumn{2}{c|}{ShanghaiTech PartA}& \multicolumn{2}{c|}{ShanghaiTech PartB} & \multicolumn{2}{c|}{SmartCity}  \\
      \hline
     Measures & MAE& MSE & MAE & MSE & MAE & MSE\\
     \hline
    YOLO9000~\cite{redmon2017cvpr} ($cs>0.05$)& 268.2 & 428.7 & 46.0& 81.6& \textbf{3.5} & \textbf{4.7} \\
    YOLO9000~\cite{redmon2017cvpr} ($cs>0.5$) & 366.8 & 513.6 & 69.0& 108.0& 5.0 & 6.0 \\
    SaCNN & \textbf{86.8} & \textbf{139.2}& \textbf{16.2}& \textbf{25.8} & 8.6& 11.6 \\
     \hline
\end{tabular}
	\caption{Comparison between YOLO9000~\cite{redmon2017cvpr} and SaCNN. $cs$ denotes the confidence score. For ShanghaiTech PartA and ParB we report the pedestrian detection results, while for SmartCity we report the head detection result.}
	\label{Tab:Detection}
\end{table*}

\subsection{Results on SmartCity}\label{Sec:SmartCity}
For SmartCity there are too few pedestrians to train a model. We employ the model trained on ShanghaiTech PartB and test it on SmartCity. Crowd scenes in ShanghaiTech PartB are outdoors and relatively sparse while in SmartCity they are both indoor and outdoor scenes with few pedestrians (7.4 on average, see Table.~\ref{Tab:Statis} and Fig.~\ref{Fig:smartcity}). Referring to the multi-task ablation test on ShanghaiTech (Sec.~\ref{Sec:ShanghaiTech}), we test both SaCNN(w/o cl) and SaCNN in Table.~\ref{Tab:SmartCity} to show how much improvement is achieved by SaCNN over SaCNN(w/o cl). By adding the relative count loss in the network, the MAE and MSE clearly drops 9.2 and 11.8 points, respectively. Fig.~\ref{Fig:smartcityres} illustrates some test examples. The density map produced by SaCNN is sparser compared to that of SaCNN(w/o cl); it is more likely to fire on the real pedestrians. While the result of SaCNN(w/o cl) reflects more subtle details of the image which are unrelated to pedestrians (\eg the third row).

In Table~\ref{Tab:SmartCity} we also test~\cite{zhang2016cvpr,sam2017arxiv}\footnote{We used the published model from~\cite{zhang2016cvpr} and trained the model for~\cite{sam2017arxiv} using their released code.}: both are trained on PartB and perform poorly on SmartCity; the MAE are 40.0 and 23.4, respectively. Our best MAE is 8.6 close to the average pedestrian number 7.4. Adding the relative count loss has significantly improved the network generalization on very sparse crowd scenes.

Overall, our results are not perfect, and we believe this makes the dataset a challenging task for future works.


%
\subsection{Regression v.s. detection}
In this section we compare our regressed-based method SaCNN with a detection-based method YOLO9000~\cite{redmon2017cvpr}. There are no bounding box annotations available for crowd counting datasets, and it is often not feasible to annotate every pedestrian head with a bounding box in dense crowds (see Fig.~\ref{Fig:detection}). Thus, we train a pedestrian detector on COCO dataset~\cite{lin2014eccv} following YOLO9000 with human annotations available. We also annotate the pedestrian head bounding boxes in COCO and train a head detector. We test both detectors on ShanghaiTech PartA and PartB, as well as SmartCity. The crowd scenes in these datasets vary from very sparse to very dense (see Table~\ref{Tab:Statis}). Since the crowd images are taken from high angles, detectors trained from COCO might not generalize well on them. We thus always report the better result over the two detectors on crowd images. Results are shown in Table~\ref{Tab:Detection}. In the detection context, a good detection is measured by the intersection-over-union (IoU) between the detected bounding box and the ground truth bounding box; the threshold for confidence score ($cs$) is set to be high, \eg $cs > 0.5$. In the crowd counting context, a good prediction is measured via the difference between the estimated count and the ground truth count; having a lower threshold \eg $cs > 0.05$ ends up with a bigger prediction.
A bigger prediction however reflects a lower MAE/MSE in the crowd counting as YOLO tends to miss small objects in general in images. We illustrate both results with YOLO9000 $cs > 0.05$ and $cs > 0.5$ in Table~\ref{Tab:Detection}: our SaCNN is significantly better than YOLO9000 on ShanghaiTech PartA and PartB; but on SmartCity it is a bit inferior to YOLO9000. Overall, we argue that the strength of our regression-based SaCNN is to provide a monolithic approach that performs close to or above the state-of-the-art on a wide range of datasets, from sparse to dense. For the very sparse case a detection-based method might be superior, but it would not perform well in the dense case. It is not clear how to combine a head detector and a density based approach, and this can be our future work.

We illustrate some examples in Fig.~\ref{Fig:detection}. Pedestrian crowds in the three samples range from very sparse to very dense. We illustrate the results of both pedestrian and head detectors. Using a small threshold for confidence score produces big prediction, but the bounding boxes are not always accurate (\eg first row: PD - $cs>0.05$). In general, YOLO9000 is good at detecting big objects in sparse crowds and is bad with small objects in dense crowds. In contrast, SaCNN can generalize very well from very spare to very dense case.


\section{Conclusion}
In this paper we propose a scale-adaptive convolutional neural network (SaCNN) to automatically estimate the density maps and pedestrian numbers of crowd images. It concatenates multiple feature maps of different scales to produce a strong scale-adaptive crowd counter for each image; it introduces a multi-task loss including a relative count loss to improve the network generalization on crowd scenes with few pedestrians. The proposed method can easily adapt to pedestrians of different scales and perspectives. Extensive experiments on standard crowd counting benchmarks demonstrate the efficiency and effectiveness of the proposed method over the state-of-the-art.

We know that the pedestrian sizes vary in according to the perspective changes in crowd images; while the multiple feature maps in SaCNN fire on pedestrians of different sizes.
In future work, we want to directly embed the perspective information as a weighting layer into SaCNN. It will produce different weights to combine the feature map outputs at every pixel depending on the perspectives.
\medskip

\noindent \textbf{Acknowledgement.} We thank Holger Caesar for his careful proofreading. 

{\small
\bibliographystyle{ieee}
\bibliography{bibtex/tencent}

\begin{thebibliography}{10}\itemsep=-1pt

\bibitem{boominathan2016mm}
L.~Boominathan, S.~S. Kruthiventi, and R.~V. Babu.
\newblock Crowdnet: a deep convolutional network for dense crowd counting.
\newblock In {\em ACM MM}, 2016.

\bibitem{brostow2006cvpr}
G.~J. Brostow and R.~Cipolla.
\newblock Unsupervised bayesian detection of independent motion in crowds.
\newblock In {\em CVPR}, 2006.

\bibitem{chan2008cvpr}
A.~B. Chan, Z.-S.~J. Liang, and N.~Vasconcelos.
\newblock Privacy preserving crowd monitoring: Counting people without people
  models or tracking.
\newblock In {\em CVPR}, 2008.

\bibitem{chan2009iccv}
A.~B. Chan and N.~Vasconcelos.
\newblock Bayesian poisson regression for crowd counting.
\newblock In {\em ICCV}, 2009.

\bibitem{chen2012bmvc}
K.~Chen, C.~C. Loy, S.~Gong, and T.~Xiang.
\newblock Feature mining for localised crowd counting.
\newblock In {\em BMVC}, 2012.

\bibitem{ciregan2012cvpr}
D.~Ciregan, U.~Meier, and J.~Schmidhuber.
\newblock Multi-column deep neural networks for image classification.
\newblock In {\em CVPR}, 2012.

\bibitem{ge2009cvpr}
W.~Ge and R.~T. Collins.
\newblock Marked point processes for crowd counting.
\newblock In {\em CVPR}, 2009.

\bibitem{idrees2013cvpr}
H.~Idrees, I.~Saleemi, C.~Seibert, and M.~Shah.
\newblock Multi-source multi-scale counting in extremely dense crowd images.
\newblock In {\em CVPR}, 2013.

\bibitem{idrees2015pami}
H.~Idrees, K.~Soomro, and M.~Shah.
\newblock Detecting humans in dense crowds using locally-consistent scale prior
  and global occlusion reasoning.
\newblock {\em TPAMI}, 37(10):1986--1998, 2015.

\bibitem{kong2005bmvc}
D.~Kong, D.~Gray, and H.~Tao.
\newblock Counting pedestrians in crowds using viewpoint invariant training.
\newblock In {\em BMVC}, 2005.

\bibitem{lempitsky2010nips}
V.~Lempitsky and A.~Zisserman.
\newblock Learning to count objects in images.
\newblock In {\em NIPS}, 2010.

\bibitem{li2008icpr}
M.~Li, Z.~Zhang, K.~Huang, and T.~Tan.
\newblock Estimating the number of people in crowded scenes by mid based
  foreground segmentation and head-shoulder detection.
\newblock In {\em ICPR}, 2008.

\bibitem{lin2017cvpr}
T.-Y. Lin, P.~Doll{\'a}r, R.~Girshick, K.~He, B.~Hariharan, and S.~Belongie.
\newblock Feature pyramid networks for object detection.
\newblock 2017.

\bibitem{lin2014eccv}
T.-Y. Lin, M.~Maire, S.~Belongie, J.~Hays, P.~Perona, D.~Ramanan,
  P.~Doll{\'a}r, and C.~L. Zitnick.
\newblock Microsoft coco: Common objects in context.
\newblock In {\em ECCV}.

\bibitem{liu2015iccv}
B.~Liu and N.~Vasconcelos.
\newblock Bayesian model adaptation for crowd counts.
\newblock In {\em ICCV}, 2015.

\bibitem{marana1998sibgrapi}
A.~Marana, L.~d.~F. Costa, R.~Lotufo, and S.~Velastin.
\newblock On the efficacy of texture analysis for crowd monitoring.
\newblock In {\em SIBGRAPI}, 1998.

\bibitem{noh2015iccv}
H.~Noh, S.~Hong, and B.~Han.
\newblock Learning deconvolution network for semantic segmentation.
\newblock In {\em ICCV}, 2015.

\bibitem{onoro2016eccv}
D.~Onoro-Rubio and R.~J. L{\'o}pez-Sastre.
\newblock Towards perspective-free object counting with deep learning.
\newblock In {\em ECCV}, 2016.

\bibitem{paragios2001cvpr}
N.~Paragios and V.~Ramesh.
\newblock A mrf-based approach for real-time subway monitoring.
\newblock In {\em CVPR}, 2001.

\bibitem{rabaud2006cvpr}
V.~Rabaud and S.~Belongie.
\newblock Counting crowded moving objects.
\newblock In {\em CVPR}, 2006.

\bibitem{redmon2017cvpr}
J.~Redmon and A.~Farhadi.
\newblock Yolo9000: better, faster, stronger.
\newblock In {\em CVPR}, 2017.

\bibitem{regazzoni1996sp}
C.~S. Regazzoni and A.~Tesei.
\newblock Distributed data fusion for real-time crowding estimation.
\newblock {\em Signal Processing}, 53(1):47--63, 1996.

\bibitem{ronneberger2015miccai}
O.~Ronneberger, P.~Fischer, and T.~Brox.
\newblock U-net: Convolutional networks for biomedical image segmentation.
\newblock In {\em MICCAI}, 2015.

\bibitem{ryan2009dicta}
D.~Ryan, S.~Denman, C.~Fookes, and S.~Sridharan.
\newblock Crowd counting using multiple local features.
\newblock In {\em DICTA}, 2009.

\bibitem{sam2017arxiv}
D.~B. Sam, S.~Surya, and R.~V. Babu.
\newblock Switching convolutional neural network for crowd counting.
\newblock In {\em CVPR}, 2017.

\bibitem{simonyan2015iclr}
K.~Simonyan and A.~Zisserman.
\newblock Very deep convolutional networks for large-scale image recognition.
\newblock In {\em ICLR}, 2015.

\bibitem{stewart2016cvpr}
R.~Stewart, M.~Andriluka, and A.~Y. Ng.
\newblock End-to-end people detection in crowded scenes.
\newblock In {\em CVPR}, 2016.

\bibitem{viola2003ijcv}
P.~Viola, M.~J. Jones, and D.~Snow.
\newblock Detecting pedestrians using patterns of motion and appearance.
\newblock {\em IJCV}, 63(2):153--161, 2003.

\bibitem{wang2015mm}
C.~Wang, H.~Zhang, L.~Yang, S.~Liu, and X.~Cao.
\newblock Deep people counting in extremely dense crowds.
\newblock In {\em ACM MM}, 2015.

\bibitem{wang2011cvpr}
M.~Wang and X.~Wang.
\newblock Automatic adaptation of a generic pedestrian detector to a specific
  traffic scene.
\newblock In {\em CVPR}, 2011.

\bibitem{wu2005iccv}
B.~Wu and R.~Nevatia.
\newblock Detection of multiple, partially occluded humans in a single image by
  bayesian combination of edgelet part detectors.
\newblock In {\em ICCV}, 2005.

\bibitem{zhang2015cvpr}
C.~Zhang, H.~Li, X.~Wang, and X.~Yang.
\newblock Cross-scene crowd counting via deep convolutional neural networks.
\newblock In {\em CVPR}, 2015.

\bibitem{zhang2016cvpr}
Y.~Zhang, D.~Zhou, S.~Chen, S.~Gao, and Y.~Ma.
\newblock Single-image crowd counting via multi-column convolutional neural
  network.
\newblock In {\em CVPR}, 2016.

\bibitem{zhao2008pami}
T.~Zhao, R.~Nevatia, and B.~Wu.
\newblock Segmentation and tracking of multiple humans in crowded environments.
\newblock {\em TPAMI}, 30(7):1198--1211, 2008.

\end{thebibliography}
}
\end{document}